\documentclass[runningheads]{llncs}
\usepackage[T1]{fontenc}
\usepackage{graphicx}
\usepackage{subcaption}
\setkeys{Gin}{draft=false}
\usepackage{amsmath,amssymb}
\usepackage{booktabs}
\usepackage{url}
\usepackage{pgfplots}
\pgfplotsset{compat=1.18}

\usepackage{booktabs}
\usepackage{placeins}
\usepackage{subcaption}
\usepackage[font=small]{caption}
\usepackage{tikz}
\usetikzlibrary{positioning,arrows.meta}
\usepackage{microtype}
\begin{document}
\title{MedFL-Stress: A Systematic Robustness Evaluation of Federated Brain Tumor Segmentation under Cross-Hospital MRI Appearance Shift}
\titlerunning{Stress-Testing Cross-Hospital Segmentation in FL}

\author{Kiran Naseer \and Samreen Azhar \and Naveed Anwar Butt}
\authorrunning{K. Naseer et al.}
\institute{Department of Computer Science, University of Gujrat, Gujrat, Pakistan \\
    \email{kirannaseer8@gmail.com}}

\maketitle
\begin{abstract}
Federated learning lets hospitals train segmentation models without
sharing patient data, but standard evaluation reports one number:
mean Dice across all sites. That number can look competitive while
one hospital runs far behind the rest.

Our results suggest this gap is not primarily a federation problem. Under
controlled MRI appearance shift on BraTS~2020, even a centralised
model trained on all data (single seed) shows a 0.055 best--worst client gap.
Under FedAvg, federation adds a further 0.030 on top of that. In our setup, appearance shift accounts for most of the gap and federation widens it.

We test five methods -- centralised, \textsc{FedAvg},
\textsc{FedProx}, \textsc{FedBN}, and \textsc{q-FedAvg} -- under
this protocol. q-FedAvg achieves the best worst-client Dice in both
full-data and subsampled regimes while also maintaining the highest
mean Dice among FL methods: no accuracy--fairness tradeoff in our setting, though its margin over FedAvg is not significant at five seeds ($p{=}0.436$). FedProx significantly lowers the worst client across five seeds ($p{=}0.019$). Worst-client rankings are identical across both evaluation regimes, but FedBN's worst-client spread across seeds is nearly double FedAvg's — a reliability gap that single-seed evaluation cannot see.

\textbf{MedFL-Stress} packages this stress test as a reproducible
protocol. The main practical message: report worst-client Dice and confirm rankings across evaluation regimes before drawing conclusions from any FL benchmark.

\end{abstract}

\section{Introduction}

Brain tumor segmentation from multi-modal MRI is central to treatment planning and disease monitoring. U-Net-style encoder--decoder architectures \cite{ronneberger2015unet} perform impressively on benchmarks like BraTS. But these benchmarks share an assumption: all training data can be pooled in one place.

That assumption rarely holds. Patient data is sensitive, and sharing raw MRI scans across hospitals is often barred by institutional and regulatory constraints \cite{rieke2020future}. Federated learning (FL) sidesteps this: each hospital trains locally, and only model weights -- never images -- reach a central server \cite{mcmahan2017}.

Hospitals do not scan patients the same way. Scanner vendor, field strength, and acquisition protocol all shape how an MRI looks, even for the same tumour type. These appearance-driven shifts are exactly the heterogeneity federated optimisers were not designed around. Standard practice still reports one number for the resulting model: mean Dice across all hospitals.

That number hides a lot. A hospital with clean, well-normalised scans pulls the average up; a hospital with an older scanner can be running a model that fails, and the mean will not show it. For brain tumour segmentation, an 8-point Dice spread across hospitals is substantial, and it concentrates at a single site — precisely the hospital least equipped to notice it. This is a fairness problem as much as an accuracy one ~\cite{chen2023fairness,pfohl2021fairness}.

Reading the federated medical imaging literature, we found worst-hospital performance almost never reported as a primary metric, and inter-site disparity even less so. No prior work in our survey combines a controlled heterogeneity protocol, explicit worst-hospital reporting, and disparity metrics as primary targets~\cite{sheller2020,li2021fedbn,pati2021fets,manthe2024federated,yan2025fedvck}.

We introduce MedFL-Stress to close that gap: a controlled stress-testing protocol, not a new architecture or optimiser. 

A second question follows from the first. If the gap is real, is it real every time we measure it, or just in the one run we happened to report? We repeat our evaluation across five random seeds to find out.

\textbf{Contributions.}
\begin{itemize}

\item \textbf{Appearance shift accounts for most of the gap.}
Centralised training on BraTS~2020 under H3 appearance shift still
shows a 0.055 best--worst Dice gap (single seed). Federation adds 0.030 more.
To the best of our knowledge, this decomposition has not been explicitly reported in federated brain tumour segmentation.

\item \textbf{q-FedAvg leads on worst-client and mean Dice, though not significantly.}
Among all tested FL methods, q-FedAvg achieves the highest
worst-client Dice and the highest mean Dice in both evaluation
regimes, with no accuracy--fairness tradeoff in our setting. Its margin over FedAvg is not confirmed at five seeds ($p{=}0.436$), so this is a promising point estimate, not a significant result.

\item \textbf{FedProx lowers the worst client.}
FedProx's worst-client Dice is significantly below FedAvg across
five seeds ($p{=}0.019$). Proximal regularisation, designed for
heterogeneity, may backfire when the global model is already biased
toward the clean-domain client.

\item \textbf{Rankings replicate; reliability does not.}  The worst-client ordering is identical across a single-seed full-data regime and a five-seed subsampled regime — q-FedAvg best, FedProx worst, in both regimes. What changes is reliability: FedBN's worst-client spread across seeds is nearly double FedAvg's ($\pm$0.033 vs.\ $\pm$0.018), an instability a single full-data run cannot reveal. MedFL-Stress packages this as a reproducible protocol with controlled heterogeneity, per-client metrics, and multi-regime evaluation.

\end{itemize}

\section{Related Work}
\label{sec:related}

FL enables collaborative training without raw data exchange, and has been applied to brain tumour segmentation and other medical imaging
tasks~\cite{mcmahan2017,sheller2019,sheller2020,rieke2020future}. Surveys consistently flag heterogeneity and cross-site generalisation as open problems~\cite{guan2024survey,xu2023,zhou2024}, and uneven performance across sites is increasingly framed as a fairness and safety issue~\cite{chen2023fairness,pfohl2021fairness} -- yet evaluations still default to global averages.

FedProx ~\cite{li2020fedprox} addresses heterogeneity through proximal regularisation; FedBN ~\cite{li2021fedbn} keeps batch-normalisation statistics local to each client. Both target the optimisation side of heterogeneity, not its measurement.

The closest related work is the FeTS Challenge~\cite{pati2021fets,zenk2025towards},
which visualises per-institution performance on naturally distributed
BraTS data. Manthe et al.~\cite{manthe2024federated} benchmark federated
optimisers on FeTS2022 with institution-wise Dice and significance
testing. Both differ from MedFL-Stress: heterogeneity is real and
uncontrolled rather than graded and appearance-isolated, and
worst-site performance is secondary rather than primary. Neither
repeats evaluation across seeds. To our knowledge, no prior work combines all of these properties.

\section{Method}

\subsection{Problem Setting}

We consider cross-silo federated brain tumour segmentation: $K$ hospitals, each holding a local dataset $D_k = \{(x_i^k, y_i^k)\}_{i=1}^{N_k}$ of multi-modal MRI slices and segmentation masks, collaborate without sharing raw data. All clients share label semantics but differ in appearance due to scanner, protocol, and preprocessing differences. The goal is a global model that performs well \emph{everywhere}, not just on average. Slices with no tumour present in the ground truth are excluded from both training and evaluation, so Dice is always computed on slices with a non-empty reference mask.

\subsection{Segmentation Model and Baselines}

We use a 2D U-Net (encoder--decoder with skip connections, four input
channels: T1, T1ce, T2, FLAIR), predicting Whole Tumor (WT), Tumor Core (TC), and Enhancing Tumor (ET) per the BraTS protocol, trained with $L = \tfrac{1}{2}L_{BCE} + \tfrac{1}{2}L_{Dice}$. We chose 2D over 3D for computational efficiency across the many (heterogeneity $\times$ method $\times$ seed) configurations this protocol requires; absolute Dice values are therefore not directly comparable to volumetric BraTS benchmarks, and we report only relative comparisons across our own FL methods.

We evaluate four federated methods against a centralised reference model.
\textsc{FedAvg}~\cite{mcmahan2017} aggregates client parameters by a
data-size-weighted average,
$\theta^{(t+1)} = \sum_k \frac{N_k}{\sum_j N_j}\theta_k^{(t+1)}$,
and serves as the standard baseline.
\textsc{FedProx}~\cite{li2020fedprox} adds a proximal term
$\frac{\mu}{2}\|\theta - \theta^{(t)}\|^2$ to the local objective,
constraining local updates to stay near the global model; we use
$\mu{=}0.01$.
\textsc{FedBN}~\cite{li2021fedbn} shares convolutional weights but
keeps batch-normalisation statistics local, letting each client adapt
to its own intensity distribution.
\textsc{q-FedAvg}~\cite{li2020qffl} ($q{=}1$) weights each client's
update by $F_k^q$ -- its training loss raised to the power $q$ -- so, hospitals with higher loss contribute proportionally more to
aggregation. At $q{=}0$ it reduces to \textsc{FedAvg}; higher $q$
increases fairness pressure toward the worst client. We use $q{=}1$, the literature default~\cite{li2020qffl}, selected via a pilot sweep over $q \in \{0.5, 1, 2, 5\}$.

\subsection{Controlled Heterogeneity Protocol}
\label{sec:heterogeneity_protocol}

Each of the four clients receives a fixed, domain-specific intensity transformation, chosen to span the main axes of real world MRI appearance variation~\cite{guan2024survey}: \textbf{Client 1} is the clean reference (no transformation); \textbf{Client 2} applies gamma contrast, $x \leftarrow x^\gamma$, emulating scanner-specific contrast response~\cite{perezgarcia2021torchio}; \textbf{Client 3} applies an affine scale-shift, $x \leftarrow \alpha x + \beta$, emulating gain/offset differences across acquisition protocols; \textbf{Client 4} applies additive Gaussian noise followed by spatial blur, emulating SNR and resolution differences across field strengths.

Each transform is varied across four heterogeneity levels H0--H3
(Table~\ref{tab:heterogeneity_params}), applied online during training and consistently at evaluation. H0 is the homogeneous  baseline; H3 is the strongest shift used for our main results. This isolates appearance shift from other confounds such as label distribution or dataset-size imbalance.

\begin{table}[t]
\caption{Heterogeneity protocol parameters, H0--H3 (Clients 2--4; Client 1 unmodified). See Section 3.4 for sampling details.}
\label{tab:heterogeneity_params}
\centering
\small
\begin{tabular}{llcccc}
\toprule
Client & Transform & H0 & H1 & H2 & H3 \\
\midrule
2 & Gamma $\gamma$ 
    & $1.0$ & $[0.8,\,1.2]$ & $[0.6,\,1.5]$ & $[0.5,\,2.0]$ \\
\midrule
3 & Scale $\alpha$ 
    & $1.0$ & $[0.95,\,1.05]$ & $[0.9,\,1.1]$ & $[0.8,\,1.2]$ \\
  & Shift $\beta$ 
    & $0.0$ & $[-0.03,\,0.03]$ & $[-0.07,\,0.07]$ & $[-0.1,\,0.1]$ \\
\midrule
4 & Noise $\sigma$ 
    & $0.0$ & $0.01$ & $0.03$ & $0.05$ \\
  & Blur $\kappa$ (px) 
    & $0.0$ & $1.0$ & $2.0$ & $[3,\,5]$ \\
\bottomrule
\end{tabular}
\end{table}

\subsection{Training and Evaluation Protocol}
\label{sec:training_protocol}

\textbf{Full-data protocol} (Section~\ref{sec:full-data}): all client
training slices, 10 communication rounds, 1 local epoch per round,
AdamW (lr$=2\times10^{-4}$, weight decay $10^{-5}$), FP16, single
seed per method. Fixed H3 parameters: $\gamma{=}1.5$,
$\alpha{=}0.8$, $\beta{=}0.1$, $\sigma{=}0.05$, blur $=3$px. All
methods plateau by round 8--9; rankings are stable from round 6.

\textbf{Subsampled protocol} (Section~\ref{sec:multiseed}): 200
randomly subsampled slices per client per round, 20 rounds, 5 seeds
(42, 123, 456, 789, 1001) applied identically across all four
methods. H3 transform parameters sampled per batch from the full
ranges in Table~\ref{tab:heterogeneity_params}. Absolute Dice
differs from the full-data protocol and is not directly comparable
(Section~\ref{sec:multiseed}); relative rankings and gap stability
are the target quantities.

In both settings we report mean Dice per subregion (WT, TC, ET),
worst-client Dice, best-client Dice, and the best--worst gap.

\begin{figure}[t!]
\centering
\includegraphics[width=\linewidth]{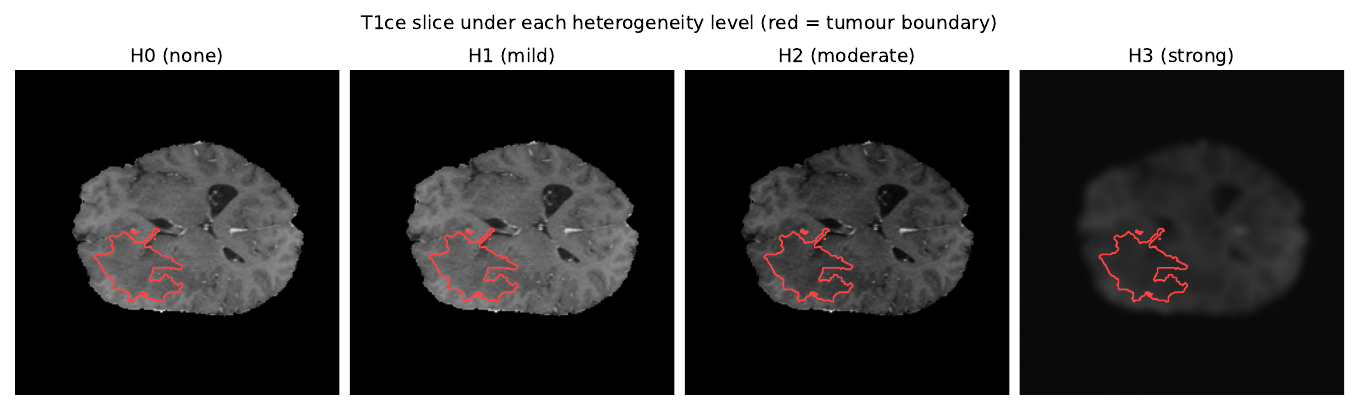}
\caption{T1ce MRI slice across heterogeneity levels H0--H3.
Red contour shows the tumour boundary (WT). H3 combines gamma
contrast, intensity scale-shift, Gaussian noise, and spatial blur --
the full appearance shift each client experiences during training
and evaluation.}
\label{fig:appearance-shift}
\end{figure}
\section{Experiments and Results}

\subsection{Dataset and Setup}

We use BraTS~2020~\cite{bakas2018,menze2015}: 369 labelled cases, four
MRI modalities (T1, T1ce, T2, FLAIR), voxel-wise WT/TC/ET annotations.
We hold out 55 cases (15\%) before any partitioning, using them solely
for evaluation. The remaining 314 cases are split into four clients
(79/79/78/78), each receiving a fixed H3 appearance transform
(Section~\ref{sec:heterogeneity_protocol}). No patient appears in more
than one client. All volumes are z-score normalised; appearance
transforms are applied on top, per client, to emulate drift that
survives standard preprocessing. Training uses 2D axial slices;
evaluation uses held-out slices under each client's own domain
transform.
\subsection{Full-Data Results}
\label{sec:full-data}

\begin{table}[t]
\centering
\caption{Full-data results under H3, single seed (42), 10 rounds,
held-out evaluation. WT/TC/ET are mean Dice per subregion across all four clients. Worst and Gap are the primary metrics; Mean is reported for completeness only.}
\label{tab:fulldata}
\small
\begin{tabular}{lccccccc}
\toprule
Method & WT & TC & ET & Mean & Worst & Best & Gap \\
\midrule
Centralised     & 0.845 & 0.598 & 0.534 & 0.659 & 0.622 & 0.678 & 0.055 \\
\midrule
FedAvg          & 0.827 & 0.587 & 0.506 & 0.640 & 0.592 & 0.676 & 0.085 \\
FedProx ($\mu{=}0.01$) & 0.785 & 0.553 & 0.478 & 0.605 & 0.560 & 0.654 & 0.094 \\
FedBN           & 0.831 & 0.572 & 0.498 & 0.634 & 0.584 & 0.665 & 0.081 \\
q-FedAvg ($q{=}1$) & 0.830 & 0.589 & 0.510 & 0.643 & 0.596 & 0.677 & 0.081 \\
\bottomrule
\end{tabular}
\end{table}

Table~\ref{tab:fulldata} shows the full picture. Two things stand out. First, the gap exists even without federation. Centralised training -- one model, all data pooled -- still shows a 0.055 best--worst gap. Appearance shift alone is enough to leave the hardest client behind; federation adds a further 0.030 on top of that baseline. The gap is not solely a federation problem. In our setup, appearance shift accounts for most of the gap, and federation widens it.

Second, mean Dice hides all of this. FedAvg reports a competitive mean of 0.640, just 0.019 below centralised. By that number, federation looks nearly free. The worst-client column shows a different picture: the weakest hospital runs 4.8 Dice points below the reported mean, and 3.0 points below the weakest hospital under centralised training.

Among the FL methods, q-FedAvg achieves the best worst-client Dice
(0.596) while also maintaining the highest mean Dice (0.643). There is no accuracy--fairness tradeoff here: weighting high-loss clients more heavily in aggregation improves the worst hospital without hurting the average. FedBN matches q-FedAvg's gap (0.081) but with a lower worst-client score (0.584). FedProx is the weakest method on both metrics -- its proximal regularisation keeps local updates close to the global model, but if that global model is already biased toward the clean client, the constraint may prevent the hardest client from adapting away from a poor starting point.

\subsection{Multi-Seed Reliability}
\label{sec:multiseed}

A single-seed result can mislead in either direction. We re-ran all
four FL methods under a matched subsampled protocol (200 slices per
client per round, 20 rounds) across five seeds (42, 123, 456, 789,
1001), holding everything else constant.

\begin{figure}[t!]
\centering
\includegraphics[width=0.92\linewidth]{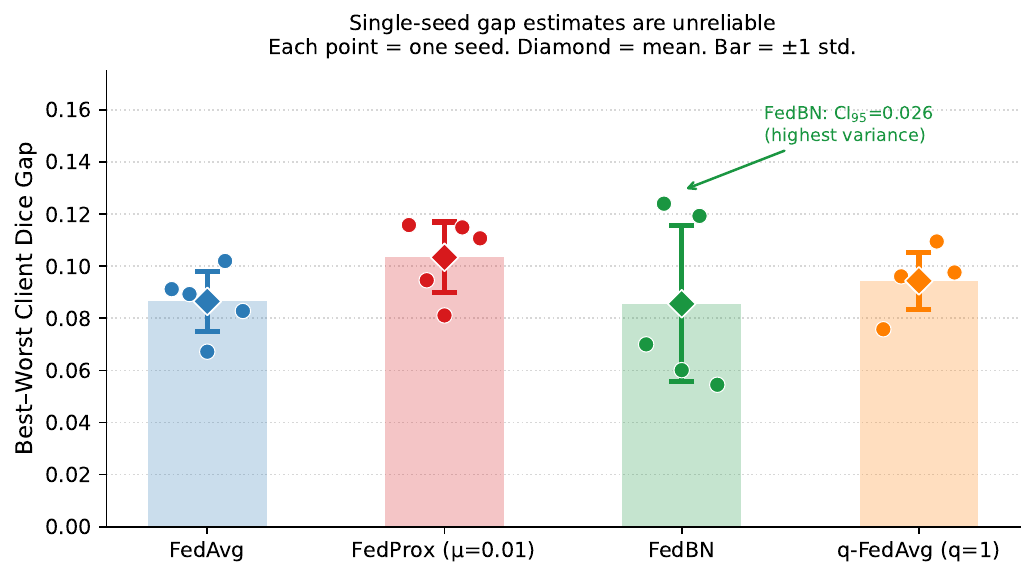}
\caption{Best--worst Dice gap across 5 seeds (subsampled protocol).
Each point is one seed; diamond = mean; bar = $\pm$1 std. The gap
is positive in every run for every method. FedBN's gap CI$_{95}$ (0.026) is roughly 2.6 times FedAvg's (0.010), consistent with a regime
interaction with local BN statistics; we did not repeat full-data FedBN across seeds, so we cannot confirm this is not also a property of FedBN at full scale.}
\label{fig:gap-instability}
\end{figure}

Three findings emerge from the five-seed evaluation (Table~\ref{tab:multiseed},
Figure~\ref{fig:gap-instability}).

\textit{The gap never closes.} Every method, every seed shows a
positive best--worst gap. No run approaches zero. In our setting the
gap persists across seeds, which suggests it is not an artefact of a
single initialisation.

\textit{FedProx lowers the worst client.} Its worst-client
Dice (0.487) is the lowest of all methods and significantly below
FedAvg ($p{=}0.019$). This is the only statistically significant result in Table~\ref{tab:multiseed} under paired t-tests across the five matched seeds.

\textit{q-FedAvg leads on worst-client Dice (0.508) and mean Dice
(0.555)}, consistent with its full-data result. The improvement over
FedAvg does not reach significance at $n{=}5$ seeds ($p{=}0.436$),
as expected given the low power of a five-sample test. The direction
matches the full-data result; the magnitude requires larger-scale confirmation.

FedBN's variability is markedly higher than the other methods
(${\pm}0.033$ on worst-client Dice vs.\ ${\pm}0.018$ for FedAvg),
visible in Figure~\ref{fig:gap-instability}. With only 200
subsampled slices per client per round, its local BN statistics are
noisily estimated -- likely a regime interaction rather than a property of
FedBN at full scale.

\begin{table}[t]
\centering
\caption{Multi-seed evaluation (5 seeds, subsampled, H3).
Mean $\pm$ 95\% CI across seeds. $p$: paired $t$-test vs.\
\textsc{FedAvg} on worst-client Dice. * $p < 0.05$.
Not directly comparable to Table~\ref{tab:fulldata}
(Section~\ref{sec:training_protocol}).}
\label{tab:multiseed}
\small
\begin{tabular}{lcccc}
\toprule
Method & Mean Dice & Worst-Client Dice & Gap & $p$ vs FedAvg \\
\midrule
FedAvg
  & $0.548 \pm 0.015$
  & $0.502 \pm 0.018$
  & $0.087 \pm 0.010$
  & --- \\
FedProx ($\mu{=}0.01$)
  & $0.541 \pm 0.015$
  & $0.487 \pm 0.018$
  & $0.103 \pm 0.012$
  & $0.019^*$ \\
FedBN
  & $0.542 \pm 0.017$
  & $0.497 \pm 0.033$
  & $0.086 \pm 0.026$
  & $0.655$ \\
q-FedAvg ($q{=}1$)
  & $0.555 \pm 0.016$
  & $0.508 \pm 0.019$
  & $0.094 \pm 0.010$
  & $0.436$ \\
\bottomrule
\end{tabular}
\end{table}

\subsection{Are Rankings Stable Across Evaluation Regimes?}
\label{sec:regime}

\begin{figure}[h]
\centering
\includegraphics[width=0.92\linewidth]{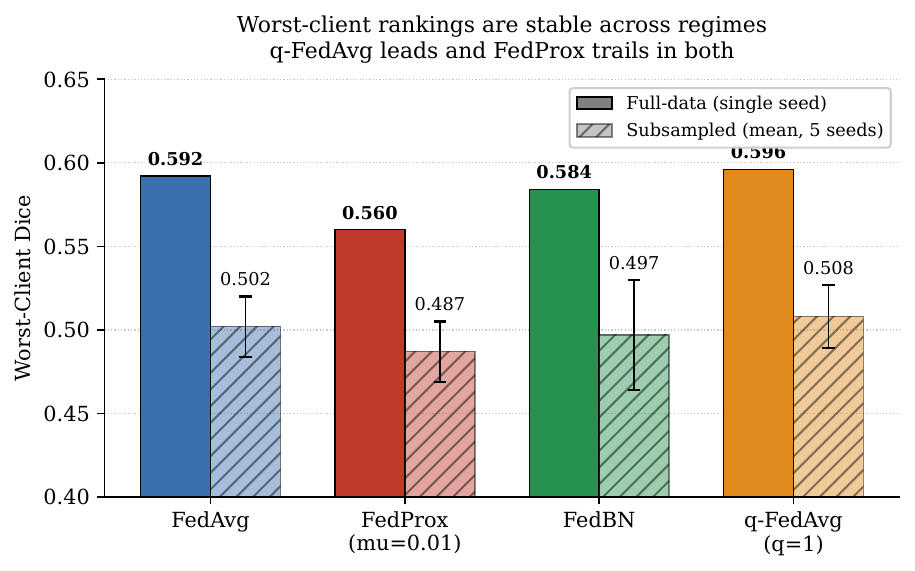}
\caption{Worst-client Dice by method and evaluation regime. Solid bars: full-data, single seed. Hatched bars: subsampled, mean over 5 seeds. The ordering is identical in both regimes — q-FedAvg first, FedProx last — so the practical recommendation does not depend on the evaluation regime.}
\label{fig:regime}
\end{figure}
Figure 3 compares worst-client Dice across the two evaluation regimes. The ordering is identical in both: q-FedAvg first, then FedAvg, then FedBN, with FedProx last. The recommendation a practitioner would draw — prefer q-FedAvg, avoid FedProx — is regime-invariant on the five-seed mean.

Multi-regime evaluation confirms the ranking rather than overturning it, but exposes what a single regime hides: the gap metric reorders across regimes while the worst-client floor does not. We therefore report worst-client Dice — a floor that ranks consistently — as the primary metric.

\section{Discussion}
\label{sec:discussion}

\textbf{The gap is real and larger than mean Dice suggests.}
FedAvg's mean Dice (0.640) sits just 0.019 below centralised --
close enough that standard reporting would call federation nearly
free. The worst client tells a different story. The best–worst gap widens from 0.055 under centralised training to 0.085 under FedAvg — a 0.030 increase — and the weakest site falls 3.0 Dice points below its centralised counterpart. In our setup, appearance shift accounts for most of the gap, and federation widens a gap that is already there.

\textbf{q-FedAvg is a promising default under appearance shift.}
It achieves the best worst-client Dice in both regimes (Table~\ref{tab:fulldata}, Table~\ref{tab:multiseed}) while maintaining the highest mean Dice among FL methods. There is no accuracy--fairness tradeoff in this setting, although its margin over FedAvg is not significant at five seeds ($p{=}0.436$).

\textbf{FedProx underperforms under appearance shift.}
Proximal regularisation was designed to control client drift under
optimisation heterogeneity. Under appearance shift, it lowers the worst client; a plausible explanation is that forcing local updates to stay near a global model already biased toward the clean domain limits how far the hard-domain client can adapt. The drop is significant across five seeds ($p{=}0.019$).

\textbf{Limitations.} Four clients, one dataset, 2D slices not
directly comparable to volumetric benchmarks. The multi-seed check
runs on a subsampled regime for tractability -- a limit of the
reliability check itself. Our transforms cover appearance variability but not annotation disagreement, demographic shift, or workflow differences. We report quantitative results only at the most severe heterogeneity level (H3); H1 and H2 are defined but not evaluated, so we cannot show how the gap scales with shift severity. We do not include a local-only baseline (each client trained on its own data alone), so we cannot say whether federation helps or harms the worst client relative to that alternative. Centralised training and full-data FedBN are each evaluated at a single seed; their point estimates (the 0.055/0.030 decomposition and FedBN's full-data worst-client Dice) are not confirmed across seeds.

\section{Conclusion}

Under controlled appearance shift, the worst-client gap is already present in centralised training (0.055, single seed), and federation widens it by a further 0.030 — in our setup, appearance shift accounts for most of the gap, with federation adding to it. Across five seeds, three results hold: the best–worst gap never closes; q-FedAvg leads on worst-client Dice while keeping the highest mean, at no accuracy cost, though its edge over FedAvg is not confirmed at five seeds (p=0.436); and FedProx significantly lowers the worst client (p=0.019). Worst-client rankings are identical across both evaluation regimes, but FedBN's worst-client spread across seeds is nearly double FedAvg's, so a single-seed study can mistake an unreliable method for a safe one.

\noindent\textbf{Code Availability.} Released upon acceptance.

\footnotesize                
\bibliographystyle{splncs04}
\bibliography{references}

\end{document}